\documentclass[letterpaper, 10pt, conference]{ieeeconf}  % Comment this line out if you need a4paper

\usepackage{wrapfig}

\usepackage{xcolor}

\IEEEoverridecommandlockouts                              % This command is only needed if 
\usepackage{amsmath}
\title{\LARGE \bf
 Impact-aware humanoid robot  motion generation with a \\quadratic optimization controller
}

\author{Yuquan Wang$^{1}$, Arnaud Tanguy, Pierre Gergondet and Abderrahmane Kheddar% <-this % stops a space
\thanks{$^{1}$
CNRS-University of Montpellier, LIRMM, Interactive Digital Humans group, Montpellier, France.
Email: \{yuquan.wang, arnaud.tanguy, pierre.gergondet, kheddar\}@lirmm.fr}
}

\usepackage{amsmath, amsthm, amsfonts}
\usepackage{amssymb}
\usepackage{graphicx}

\usepackage[font=normal,labelfont=bf]{caption}
\usepackage{subcaption}
\usepackage{hyperref}
\newcommand{\bs}{\boldsymbol}

\newtheorem{remark}{Remark}[section]

\begin{document}

\maketitle
\thispagestyle{empty}
\pagestyle{empty}

%%%%%%%%%%%%%%%%%%%%%%%%%%%%%%%%%%%%%%%%%%%%%%%%%%%%%%%%%%%%%%%%%%%%%%%%%%%%%%%%
\begin{abstract}
Impact-aware tasks (i.e. on purpose impacts) are not handled in multi-objective whole-body controllers of humanoid robots. This leads to the fact that a humanoid robot typically operates at near-zero velocity to interact with the external environment. We explicitly investigate the propagation of the impact-induced velocity and torque jumps along the structure linkage and propose a set of constraints that always satisfy the hardware limits, sustain already established contacts, and the stability measure, i.e. the zero moment point condition. Without assumptions on the impact location or timing, our proposed controller enables humanoid robots to generate non-zero contact velocity without breaking the established contacts or falling. The novelty of our approach lies in building on existing continuous dynamics whole body multi-objective controller without the need of reset-maps or hybrid control.
\end{abstract}

%%%%%%%%%%%%%%%%%%%%%%%%%%%%%%%%%%%%%%%%%%%%%%%%%%%%%%%%%%%%%%%%%%%%%%%%%%%%%%%%
\section{Introduction}
\begin{figure}[!htp]
 \centering
 \includegraphics[width=\columnwidth]{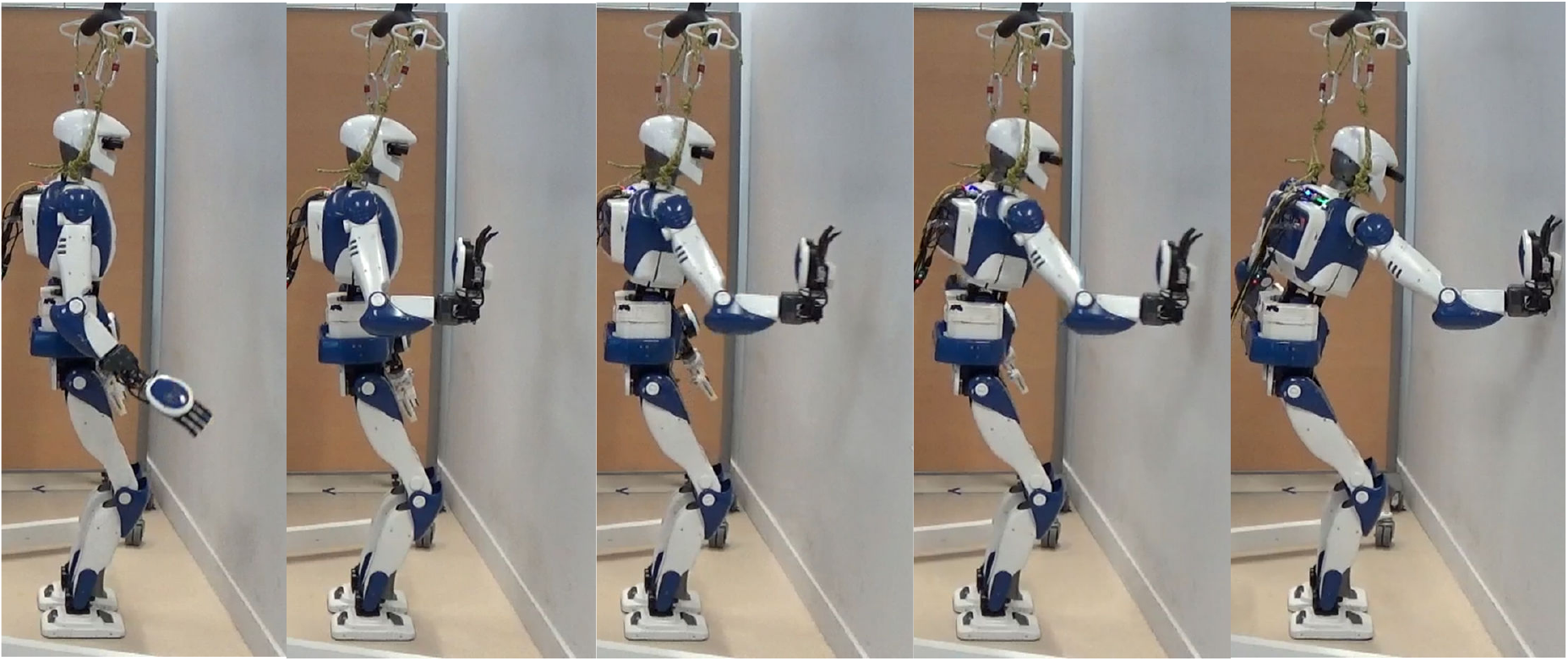}
 \caption{Snapshots of the HRP-4 robot pushing a concrete wall. The contact velocity is $0.35~m/s$ at the impact time which is determined from force sensor readings. }
\label{fig:experiment-snapshots}
\vspace{-2mm}
\end{figure}

Advanced humanoids capabilities such as walking and manipulation improved substantially in recent years. Yet, when it comes into general-purpose loco-manipulation, humanoid robots fear impacts similarly to most existing robots. Dealing with task-aware impacts --e.g. on purpose impact tasks such as pushing (see Fig.~\ref{fig:experiment-snapshots}) and even hammering or landing at jumps... or non-desired impacts --e.g. those consequent to falls, requires capabilities in both the hardware design and the controller aspects.

Impacts last a very short of time~\cite{zheng1985mathematical} (in theory, it is instantaneous), in which a considerable amount of energy is propagated through the structure and linkage of the humanoid robot and could potentially result in (i) hardware damage, and (ii) a jump in some or all unilateral contacts that existed before impact. A large part of handling impact must be tackled from a hardware perspective that we do not address in this paper. We rather assume that we possess knowledge on tolerable impact bounds, that the linkage mechanics, actuators, and electronics can absorb without damage. Indeed,  no controller can deal with any strategy at the very impact-instant: the energy shall simply be absorbed by the hardware. However, a controller can be designed to act before and after the impact, most robustly and stably.

% \begin{figure}[!htp]
% \vspace{-2mm}
% \centering
% \includegraphics[width=\columnwidth, height=3.6cm]{figure/experiment-sep-06/box-grabbing-crop.pdf}
% \caption{The HRP-4 robot grabs a box using swift motion where the contact velocity for two palms are $0.15~m/s$. }
% \label{fig:snatch_box}
% \vspace{-5mm}
% \end{figure}

It is not easy to design an impact-aware whole body controller that can  achieve on-purpose impact tasks due to the following facts:\\
{\bf(1)} impacts induce jumps in part of the robot state: that is, abrupt changes of the joint velocities, torques, and --in the case of humanoids, unilaterally established contact forces;\\
{\bf(2)} Due to the jumps, the robot dynamics (equations of motions) are different and a reset map is needed;\\
{\bf(3)} the difficulty (if not the impossibility) to know precisely some pertinent parameters, such as the environment stiffness, the coefficient of restitution, the impact localization on the robot (and the environment), the contact normal, and the exact impact time;

These parameters are pertinent to model impact dynamics and their uncertainty might cause undesired post-impact status, e.g. rebounce or sliding. Therefore a common practice is to set and release a contact at near-zero velocity to ensure a smooth contact transition without invoking impacts. 

We propose to overcome these limitations by integrating the impact dynamics model into our whole-body multi-objective continuous dynamics controller (and not specifically design a dedicated controller to handle task-aware impact). This choice is very important and constitutes the main novelty and the most appealing aspect of our approach w.r.t. e.g. existing task-specific controllers, reset map controllers or hybrid controllers, etc. The main idea is to guarantee, through considering upper-bounds, the worst-case impact situations such that the robot motion is robust to an impact whose exact timing, location, and other pertinent parameters that might be not known exactly.

Our whole-body robot controller is formulated as quadratic programming in the task space~\cite{bouyarmane2019tro}. We have demonstrated our controller with very complex multi-objective dynamic operations and embed already visual servoing, force control, set-point and trajectory tracking tasks under various types of constraints such as joint limits, collision avoidance, etc. Our main goal is to extend this controller with multi-purpose impact-aware tasks with minimal structural changes and if possible, no particular switching or \emph{if-then-elses}, i.e. only by designing additional impact-aware constraints and tasks that can be added or removed at will and on-purpose. We stress on the importance of this choice because it allows having an enhanced integrated multi-purpose control framework.  

In our previous work~\cite{wang2019rss}, we show that for a fixed-based robot, hardware limitations in terms of max allowable impact can be easily integrated as additional constraints in our controller. Yet, such constraints do not prohibit jumps in the joint velocities and torques that could make the controller computation fails or diverge in a closed-loop scheme. This is because, right after the impact, the QP solver might start from an unfeasible constraints set. These facts are obviously found in humanoid robots too. In humanoids, we also have unilateral contacts setting and a floating (under-actuated) base: not only the joint velocities and torques undergo an abrupt more or less substantial change, but so does each contact forces. By applying impact dynamics analysis to the operational space equations of motion, one can model the propagation of the state jumps between the end-effectors, see Sec.~\ref{sec:proposed_qp}. 
% As an example shown in , the simulated HRP-4 robot lost the ground contact due to the impact on the palm.

Moreover, the dynamic balance of humanoid robots --eventually through the Zero Moment Point (ZMP), under impacts has been investigated for planning purposes in specific tasks~\cite{tsujita2008iros, konno2011impact}. A multi-objective controller that fulfills the dynamic balance constraint under impacts is, to our best knowledge, missing, see Sec.~\ref{sec:related_work}. Indeed, sustaining at best prior contacts and balance during impacts is a fundamental issue that is not yet explicitly addressed in any existing QP controller frameworks. We highlight this gap in Sec.~\ref{sec:common_qp} based on the analysis performed on a state-of-the-art QP controller.
We analytically derive inequality constraints to generate feasible robot motion such that the impact-induced state jumps will not break the contact and balance conditions regardless of the impulsive forces. 

We assess our impact-aware multi-objective QP controller using an HRP-4 humanoid robot that impacts a fixed concrete wall without knowing exactly its location in Sec.~\ref{sec:experiments}.
% achieved experiments using the  HRP-4 humanoid robot presented in Sec.~\ref{sec:experiments}. The first experiment consists in 
% The second experiment consists in swiftly grab a box from one location and put it down fastly. 
The impact-awareness
% constraints 
% We propose a modified formulation of the QP controller constraints to
enables
% generating 
% impact friendly robot motion based on the predicted post-impact effects, i.e. the propagation of the joint velocity jumps, impulsive forces and its influence to the existing unilateral contacts and ZMP conditions. Now
a humanoid robot to apply impacts without stopping or reducing speed.

\section{Related work}
\label{sec:related_work}

Impact duration analysis in~\cite{tsujita2008iros,pashah2008prediction} revealed that even for low-velocity, the duration of an impact is typical of milliseconds order or less. In such a short period it is difficult to devise an efficient controller that prevents hardware to be hindered to some extent. For instance, even if a variable stiffness actuator lower damage risks at impacts, it needs more than 10~ms to generate the joint torque that can counterbalance the impulsive torques~\cite{haddadin2009requirements}. Therefore, our controller doesn't consider and is independent of impact timing.

The discrete impact dynamics model has been introduced into robotics since late 1980~\cite{zheng1985mathematical}. Yet, 
more refined physics laws for multiple contacts and impacts are not known for inelastic impacts until around 2010~\cite{stewart2000rigid, featherstone2014rigid}. Recently a flying object batting example is developed in~\cite{jia2019ijrr}, where a closed-form 2D impact dynamics model is used to generate desired impulsive forces. However in the 3D cases, the closed-form solution is only available if we can control the initial sliding direction to the invariant subset~\cite{jia2017analysis}. Thus we restrict ourselves to the impact models based on algebraic equations~\cite{zheng1985mathematical} that have been successfully applied in multiple scenarios~\cite{tsujita2008iros, konno2011impact, rijnen2017control}. To our best knowledge, on-purpose impact tasks are studied only in very few work e.g. in~\cite{konno2011impact} for specific tasks. However, their controller doesn't account for uncertainties in most impact parameters, it is based on a non-linear optimization for planning, and doesn't account explicitly for constraints in the closed-loop motion. Our aim is to extend state-of-the-art task space multi-objectives and multi-sensory whole-body control framework formulated as QP to encompass impact tasks.

Impact dynamics is not well exploited by state-of-the-art practical control strategies. Most impact stabilization papers require \emph{flexible} models with regularization, e.g. the mass-spring-damper~\cite{hu2007energy, stanisic2012adjusting, heck2016guaranteeing}. Considered as a transient behavior, impact dynamics is used in stability analysis~\cite{pagilla2001stable} rather than in explicit control design. Ths exception is impact models integrated explicitly in an humanoid walking controller based on hybrid zero dynamics, e.g.~\cite{hurmuzlu2004automatica,vanzutven2010simpar,grizzle2010sncs,hereid2018tro}. Yet these approaches result in a hybrid control scheme that we aim to avoid. 

In our recent work~\cite{wang2019rss}, we use a task-space force controller~\cite{bouyarmane2019tro} to inhibit oscillations and use explicit upper bounds on the impact-induced state jumps to account for hardware limitations, e.g. bounds on impulsive forces and velocity jumps. To the best of our knowledge, our work is the first to embed high-velocity contact-task in the QP formulation that is safe to deal with impacts, while accounting for both hardware limitations and controller feasibility. However, our previous study was achieved for fixed-based robots. When humanoids are to be used, the under-actuated floating-base and balance must be taken into account. Hence, we analytically derive constraints that sustain prior-to-impact unilateral contacts and whole-body balance conditions under impacts that are seamlessly integrated in the continuous dynamic domain multi-objective QP controller~\cite{bouyarmane2019tro}.  

% Sustaining unilateral contacts is necessary for a stable robot/environment physical interaction. State-of-the-art approaches achieve near-zero relative motion between the end-effector and the contact surfaces, keep the CoP inside the contact area and restrict the contact force within the Coulomb friction cone to not slide. Representing requirements from these three aspects by a Gravito-Inertial Wrench Cone (GIWC)~\cite{caron2015leveraging}, we obtain the maximum perturbations that the robot can resist at a given configuration, and/or the maximum interaction force that the robot can generate at a given posture. Recently the GIWC cone is introduced to the instantaneous control of robot posture~\cite{abi2019torque}. However impact dynamics, i.e. the impact-induced jumps in velocity, torques and forces are not considered by controllers, e.g. \cite{caron2015leveraging, abi2019torque}

The ZMP is widely used as a balance criterion for biped walking~\cite{kajita2014introduction}, and recently extended to a multi-contact setting in~\cite{caron2016tro}. For trajectory planning tasks that require large impulsive forces, e.g. a nailing task, such as in~\cite{tsujita2008iros} and a wooden piece breaking task in~\cite{konno2011impact}, ZMP is used to analyze the stability of each robot configuration instance. Introducing the impact-robust ZMP constraint into the QP controller allows more reliable and robust motions generation.

\section{Continuous Time-Domain QP Formulation}
\label{sec:common_qp}
The detailed QP formulation can be found in~\cite{bouyarmane2019tro}. Here we focus on the most pertinent parts we use, i.e. the contact and ZMP constraints in Sec.~\ref{sec:contact} and Sec.~\ref{sec:zmp} respectively. We then summarize the usual form of the QP controller in Sec.~\ref{sec:generic_qp} to mathematically highlight why state-of-the-art QP controllers could become infeasible. 

\subsection{Contact Constraint}
\label{sec:contact}

\subsubsection{Geometric Constraint}
For each contact of the robot with its surrounding, differentiation of the kinematics model leads to $J\ddot{\bs{q}} + \dot{J}\dot{\bs{q}} = 0$, where $J(\bs{q})$ is the robot contact Jacobian and $\bs{q}$ denotes the generalized coordinate of the robot. We restrict zero relative motion at the contact by:
\begin{equation}
\label{eq:contact_geo_constraint}
J\ddot{\bs{q}} + \dot{J}\dot{\bs{q}} = -\frac{\bs{v}}{\Delta t},
\end{equation}where $\bs{v}$ denotes the actual robot contact-point velocity, and $\Delta t$ denotes the sampling period.

\subsubsection{Center of Pressure Constraint}
Assuming we have a given number of adjacent contact points forming a closed convex contact planar surface $\cal S$ with a single contact normal $n$. In view of the local external force $\bs{f}$ and moment $\bs{\tau}$, the center of pressure (CoP) is given as: 
$
\bs{p}_x = -\frac{\bs{\tau}_y}{\bs{f}_n},  \quad \bs{p}_y = \frac{\bs{\tau}_x}{\bs{f}_n}.
$
% which implies that due to the cross product in \eqref{eq:ground_reaction_moment}, we can not simultaneously regulate the contact force $\bs{f}_{\bs{e}}$ and the COP location. 
% that is able to regulate the desired ground reaction forces of the individual foot contact, is a suboptimal but feasible solution. 
% It sequentially solves for $\bs{f}_{\bs{e}}$, and then $\bs{\tau}_{\bs{e}}$ and location of COP. For simplicity, we take $\bs{f}_z$ from the last step as an known variable and regulate $\bs{\tau}_x$ and $\bs{\tau}_y$ to adjust the COP position $\bs{p}_x$ and $\bs{p}_y$.
$x$ and $y$ are the contact tangent space components, and 
\begin{equation}
[\bs{p}_x, \bs{p}_y] \in {\cal S}
\label{eq:CoP_constriants_ini}
\end{equation}
%Assuming the contact area is given as
% with two inequality constraints, i.e. 
%$\bs{p}_x \in [\underline{\bs{p}}_x, \bar{\bs{p}}_x]$ and $\bs{p}_y \in [\underline{\bs{p}}_y, \bar{\bs{p}}_y]$, which are equivalent to the following inequalities:  
% we can use the following constraint on $\bs{\tau}_z$ to keep  the COP position bounded: 
% we can fix $\bs{f}_z$ from the QP controller and 
% In order to facilitate the discussion, we use a predefined COP position $\bs{p}^*$ and external force   $\bs{f}^{*}_{\bs{e}}$ to infer the external moment and $\bs{\tau}_{\bs{e}}$: 
% \begin{equation}
% \label{eq:CoP_constriants}
% \bs{\tau}^*_{\bs{e}} = -\bs{p} \times \bs{f}_{\bs{e}} = \bs{f}^*_{\bs{e}} \times \bs{p}^*
% \end{equation}
% where  $\bs{f}*_{\bs{e}}$ and $\bs{\tau}*_{\bs{e}}$ can be achieved as an objective of the QP controller. 
%\begin{equation}
%\label{eq:CoP_constriants_ini}
%\begin{aligned}
%\underline{\bs{p}}_x &\leq -\frac{\bs{\tau}_y}{\bs{f}_n}  ~\leq  \bar{\bs{p}}_x \\
%\underline{\bs{p}}_y &\leq  ~\frac{\bs{\tau}_x}{\bs{f}_n} ~\leq  \bar{\bs{p}}_y
%\end{aligned} \Rightarrow 
%\begin{aligned}
%- \bs{\tau}_y - \bar{\bs{p}}_x\bs{f}_n & \leq 0\\
% \bs{\tau}_y + \underline{\bs{p}}_x\bs{f}_n & \leq 0 \\
%\bs{\tau}_x - \bar{\bs{p}}_y\bs{f}_n & \leq 0\\
%-\bs{\tau}_x + \underline{\bs{p}}_y\bs{f}_n & \leq 0
%\end{aligned}.
%\end{equation}
As long as the contact persists, i.e. $\bs{f}_n > 0$, the constraints~\eqref{eq:CoP_constriants_ini} are always non-singular.
We can reformulate the CoP constraint~\eqref{eq:CoP_constriants_ini} in a matrix form, that is: 
\begin{equation}
\label{eq:CoP_constriants}
A_c \bs{f}\leq \bs{0}.
\end{equation}
%where $A_{c} \in \mathbb{R}^{4\times 6}$ is given by: 
%$$
%  \begin{bmatrix}
%    0& 0& -\bar{\bs{p}}_x& 0& -1& 0\\
%    0& 0& \underline{\bs{p}}_x& 0& 1& 0\\
%    0& 0& -\bar{\bs{p}}_y& 1& 0& 0\\
%    0& 0& \underline{\bs{p}}_y& -1& 0& 0
%  \end{bmatrix}.
%$$

\subsection{Bounded ZMP}
\label{sec:zmp}
If using the zero moment point (ZMP) as the dynamic equilibrium criteria, the ZMP point shall be inside the support polygon having normal $n$ (in co-planar contacts): $\bs{z} \in \mathcal{S}$. When $n = [0,0,1]^\top$ (opposite to the gravity direction), the ZMP expresses as:
\begin{equation}
\label{eq:ZMP_ground_reaction}
\bs{z}_x    = 
-\frac{\sum \bs{\tau}_{y}}{\sum \bs{f}_n},\quad 
\bs{z}_y    =
\frac{ \sum \bs{\tau}_{x} }{\sum \bs{f}_n}.
\end{equation}
We assume that the support polygon is convex and the half-plane representation  $\bs{A}_{\bs{x}}, \bs{A}_{\bs{y}}, \bs{B} \in \mathbb{R}^{n \times 1}$ is available: 
$$
\begin{bmatrix}
\bs{A}_{\bs{x}} & \bs{A}_{\bs{y}} 
\end{bmatrix}
\begin{bmatrix}
\bs{z}_x\\
\bs{z}_y
\end{bmatrix}
\leq 
\bs{B}.
$$
Substituting $\bs{z}_x,\bs{z}_y$ defined by \eqref{eq:ZMP_ground_reaction}, we can obtain the following constraint
on the external wrenches $\sum \bs{F}$:
\begin{equation}
\label{eq:ZMP_constraints}
\underbrace{
\begin{bmatrix}
\bs{A}_{\bs{y}} & - \bs{A}_{\bs{x}} & \bs{0}  & \bs{0}  & \bs{0}  & -\bs{B}
\end{bmatrix}}_{A_{\text{Z}}}
\sum \bs{F}
\leq 0.
\end{equation}

\subsection{QP controller for a humanoid robot}
\label{sec:generic_qp}

Our QP controller is built from desired task objectives (that shall be met at best in the QP cost function) and gathers desired task constraints (that shall be met strictly, as part of the QP constraints). Thus, in the continuous time-domain, our QP controller for a humanoid robot is enhanced by the previous constraints, in plus of the common usual ones such as joint limits, collision avoidance, torque limits... that we do not mention:
\begin{equation}
\label{eq:humanoids_qp}  
\begin{aligned}
  \min_{
\bs{x}: (\ddot{\bs{q}}, \bs{f}_{\lambda})} \quad & \sum_{i \in \mathcal{I}_{o}}  w_i \|\bs{e}_i (\bs{x}) \|^2   
  \\
  \mbox{s.t.}\quad & 
   \text{Common usual constraints},\\
  \quad& \text{Contact constraints: }~\eqref{eq:contact_geo_constraint},~\eqref{eq:CoP_constriants},  \\
  \quad& \text{Bounded ZMP: }~\eqref{eq:ZMP_constraints}, 
\end{aligned}
\end{equation}
where the set  $\mathcal{I}_{o}$ can include any task, e.g. motion tasks, impedance tasks and so on; $\bs{e} (\bs{x})$ denotes the task error function weighted by $w_i$. $\bs{e} (\bs{x})$ is linear in terms of the decision variable $\ddot{\bs q}$ and discretized friction cone contact forces $\bs{f}_{\lambda}$, see~\cite{bouyarmane2019tro} for more details,
% but also contains non-linear terms in function of $\dot{\bs q}$ and $\bs q$;
so are all the constraints of the QP.

Impacts result in instantaneous jumps of the joint velocities~$\dot{\bs{q}}$, joint torques, and contact forces~$\bs{f}$, which are present in the constraints of the QP controller~\eqref{eq:humanoids_qp}.
% and updated continuously from the robot state estimators or sensors measurements. In this case, the feasibility of~\eqref{eq:humanoids_qp} is not guaranteed.
Indeed, such an abrupt jump could result in a not feasible QP for the next control iteration and no command can be issued
% If the QP is still feasible, then the generated command could result in an divergent or unstable behavior
as exemplified in~\cite{wang2019rss}. For humanoids, it can also result on falls.
 % (see, Fig.~\ref{fig:failure_humanoids}), see also Sec.~\ref{sec:impact_dynamics}. 

% Moreover, unilateral contacts do not provide pulling forces. Impulsive forces propagate in the humanoid structure through the links and linkages and may break some or even all existing contacts. The joint velocity difference $\Delta \bs{\dot{q}}$, between post- and pre-impact (that we call jump), need to fulfill the post-impact contact constraints: 
% \begin{equation}
% \label{eq:post_impact_geometry_dynamics}
% J \Delta \dot{\bs{q}} = 0,   
% \end{equation}
% which is obtained from explicitly writing that the pre- and post-impact contact positions are unchanged at the impact time even if the velocities are different, e.g. see 
% \cite{grizzle2010sncs}. If there were $m$ contacts, the equality constraint~\eqref{eq:post_impact_geometry_dynamics} strictly
% restrict the joint velocity jump $\Delta \dot{\bs{q}}$ in the intersection of the null space of the $6m$ rows. The equality constraint~\eqref{eq:post_impact_geometry_dynamics} is conservative in practice. Feedback from the real robot states includes sensor noise, numerical errors or computation latency, which can easily lead to an infeasible solution.

%Thus extra efforts are needed to guarantee the hardware limitations, to sustain established contacts, and the ZMP constraint when an impact occur at known or unknown location and time.

\section{Proposed QP controller}
\label{sec:proposed_qp}

We present the estimation of impulse propagation in Sec.~\ref{sec:impact_dynamics} and \ref{sec:estimation}. The latter is used to explicitly derive the influence of the impact on the hardware limits (Sec.~\ref{sec:impact_hardware}), contact forces (Sec.~\ref{sec:impact_contact_maintenance}), and balance constraints (Sec.~\ref{sec:impact_ZMP}). We integrate impact-aware constraints in our multi-objective whole-body controller in Sec.~\ref{sec:impact_qp_controller}.

\subsection{Impact dynamics}
\label{sec:impact_dynamics}
%\label{sec:equations_of_motions}
Let $\vec{\bs{n}}$ be the impact surface normal, and $c_{\text{r}}$ the coefficient of restitution  (we discuss how these parameters are obtained later in Sec.~\ref{sec:parameters}). Projecting the 
pre-impact end-effector velocity $\dot{\bs{x}}^{-}$ along $\vec{\bs{n}}$, 
we can predict the post-impact end-effector velocity $\dot{\bs{x}}^{+}$ as
$$
\dot{\bs{x}}^{+} = -c_{\text{r}}P_{\vec{\bs{n}}}\dot{\bs{x}}^{-} + (I - P_{\vec{\bs{n}}})\dot{\bs{x}}^{-},
$$
where $P_{\vec{\bs{n}}}=\vec{\bs{n}}\vec{\bs{n}}^{\top}$. The end-effector velocity jump is defined as:
\begin{equation*}
\label{eq:ee_jump}
\Delta \dot{\bs{x}} = \dot{\bs{x}}^{+} - \dot{\bs{x}}^{-} =  \underbrace{-(1 + c_{\text{r}})P_{\vec{\bs{n}}} }_{P_{\Delta}} \dot{\bs{x}}^{-}. 
\end{equation*}
Thus at time step $k$, we predict the end-effector velocity jump $\Delta \dot{\bs{x}}_{k+1}$ as: 
\begin{equation}
\label{eq:delta_ee_v_old}
\Delta \dot{\bs{x}}_{k+1}
= P_{\Delta} \dot{\bs{x}}^{-}_{k+1},
\end{equation}
where $ \dot{\bs{x}}^{-}_{k+1} = \bs{J}_{\dot{\bs{x}}^{-}} \dot{\bs{q}}_{k+1}$,  $\bs{J}_{\dot{\bs{x}}^{-}} = \bs{J}_{k+1} = \bs{J}_k + \dot{\bs{J}}_k \Delta t$  and $\dot{\bs{q}}_{k+1}=\dot{\bs{q}}_k + \ddot{\bs{q}}_k\Delta t$. 
The $\dot{\bs{q}}_{k}$ is obtained from the robot current state and $\Delta t$ denotes the sampling period. We re-write \eqref{eq:delta_ee_v_old} as a
function of  the optimization variable $\ddot{\bs{q}}_k$:
\begin{equation}
\label{eq:delta_ee_v}
\Delta \dot{\bs{x}}_{k+1} = P_{\Delta}(\bs{J}_k\Delta t \ddot{\bs{q}}_k + \dot{\bs{J}}_k\Delta t^2 \ddot{\bs{q}}_k + \bs{J}_k\dot{\bs{q}}_k+\dot{\bs{J}}_k\dot{\bs{q}}_k\Delta t),
\end{equation}
where we can neglect the term $\dot{\bs{J}}_k\Delta t^2 \ddot{\bs{q}}_k\approx 0$ as $\Delta t \leq 5$~ms. % The calculation of the Jacobian time-derivative alone is computationally demanding. All other terms' fast computations are readily available as part of the QP control framework.  

% Now we can use \eqref{eq:delta_ee_v} as an additional QP task (constraint) to restrict the jump in the velocity. These jump predictions can be eliminated by upper-bound if known, or kept as additional QP decision variables otherwise.
% Given   $\bar{J}_m$ and $\delta \bs{x}_m(t_{k+1})$, we 
% can predict the joint velocity jump 
% $\delta \dot{\bs{q}}(t_{k+1})$: 
% \begin{equation}
% \label{eq:joint_velocity_jump}
% \delta \dot{\bs{q}}(t_{k+1}) = \bar{J}_m \delta \bs{x}_m(t_{k+1}) 
%  =\bar{J}_mP_{\delta}J_m \left(\dot{\bs{q}}(t_{k}) + \ddot{\bs{q}}(t_{k}) \Delta t\right).
% \end{equation}

\subsection{Impulse prediction}
\label{sec:estimation}
% If a robot has sustained contact when the impact happens, 

Let us consider a humanoid robot with $n$ DoF and $m$  end-effectors with established contacts. The impact is about to happen at another end-effector $m+1$ (e.g. $m=2$ in the case where two feet are in contact with the ground, one gripper is free and the other is about to achieve a desired impact with the wall). We not only need to predict the impulse $I_{m+1}$ but also how it propagates along the kinematic tree to any of the previously defined $m$ task effectors.
Namely, in addition to  $I_{m+1}$, we need to predict the propagated impulses $I_{i}$ for $i=1\cdots m$ and the impact-induced joint velocity jumps of all the kinematic branches $\Delta \dot{\bs{q}} = \dot{\bs{q}}^{+} - \dot{\bs{q}}^{-}$.

Let $\bs{x} = [\bs{x}^\top_1, \ldots, \bs{x}^\top_{m+1}]^\top \in \mathbb{R}^{3(m+1)}$ the end-effectors coordinates and associated Jacobians $J =  [J^\top_1, \ldots, J^\top_{m+1}]^\top\in \mathbb{R}^{3(m+1) \times n}$, 
we use the operational space dynamics 
(or equivalently the \emph{articulated-body inertia} presented in Sec.~7.1 of the book by Featherstone~\cite{featherstone2014rigid})  
to characterize the impulse propagation 
% operational space equations of motion to analyze the relation 
between $I_{m+1}$ and  $I_{i}$ for $i=1,\ldots, m$:
\begin{equation}
\label{eq:robot_dynamics_operational_space}
% \Lambda(\bs{q}) \ddot{\bs{x}}  +  \bs{\mu} (\bs{q}, \dot{\bs{q}}) + \bs{\rho}(\bs{q}) = \bs{f}, 
\ddot{\bs{x}} = \Lambda(\bs{q})^{-1}\bs{f} + \bs{\beta}
\end{equation}
where  
$\bs{f} \in \mathbb{R}^{3(m+1) }$ denotes all the external contact forces and the impulsive force, 
the inverse operational space inertial matrix $\Lambda(\bs{q})^{-1} \in \mathbb{R}^{3(m+1) \times 3(m+1)}$ is defined as:
$
\Lambda(\bs{q})^{-1} = J M^{-1}J^\top.
$ The remaining acceleration bias $\bs{\beta}$ that we do not use, are defined in~\cite{featherstone2010exploiting}. We can compute a first-order approximation of predicted $\Lambda(\bs{q}_{k+1})^{-1}$ as follows:
\begin{equation}
(J_k+\Delta t \dot{J}_k)(M_k+\Delta t \dot{M}_k)^{-1}(J_k^T+\Delta t \dot{J}_k^T)
\end{equation} where $\dot{M}_k = C_k+C_k^T$ computation is readily available in the QP control framework. Integrating the equations of motion  \eqref{eq:robot_dynamics_operational_space} over the impact duration $\delta t$ and expanding the inverse operational space inertia matrix $\Lambda^{-1}$, we can obtain 
\begin{equation*}
  \label{eq:impact_dynamics_general}
\begin{bmatrix}
  \Delta \dot{\bs{x}}_1 \\
  \Delta \dot{\bs{x}}_2 \\
  \vdots \\
  \Delta \dot{\bs{x}}_{m+1}
\end{bmatrix}  
=   
\begin{bmatrix}
\Lambda^{-1}_{11} & \ldots &  \Lambda^{-1}_{1(m+1)}\\
\Lambda^{-1}_{21}  & \ldots & \Lambda^{-1}_{2(m+1)} \\
\vdots &  \ddots & \vdots \\
\Lambda^{-1}_{(m+1)1} & \ldots & \Lambda^{-1}_{(m+1)(m+1)}  
\end{bmatrix}
  \begin{bmatrix}
    I_1\\
    I_2\\
    \vdots\\
    I_{m+1}
  \end{bmatrix}
\end{equation*}
where the  inverse inertial matrix $\Lambda^{-1}_{ij}$ relates the external impulse 
$
I_j = \int \bs{f}_j \delta t,
$ acting on the $j$-th end-effector to the $i$-th end-effector velocity jump $\delta \dot{\bs{x}}_i$. In a compact form we have:
\begin{equation}
  \label{eq:impact_dynamics_general_middle}
\Delta \dot{\bs{x}} = \Lambda^{-1} I
% \begin{bmatrix}
%   \delta \dot{\bs{x}}_1 \\
%   \delta \dot{\bs{x}}_2 \\
%   \vdots \\
%   \delta \dot{\bs{x}}_{m+1}
% \end{bmatrix}  
% =   
% \Lambda^{-1}
% \begin{bmatrix}
%     I_1\\
%     I_2\\
%     \vdots\\
%     I_{m+1}
%   \end{bmatrix}.
\end{equation}
% which relates the impulses to be estimated to the 
% intermediate variables, i.e. the end-effector velocity jumps.
% where $I_{m+1}$ denotes the impulse due to the di
 % $I_i$ for $i=1,\ldots, m$ are pro
% Considering the $m+1$ end-effector is about to apply an impact, 
% Defining the impact-induced joint velocity jump $\delta \dot{\bs{q}} = \dot{\bs{q}}^{+} - \dot{\bs{q}}^{-}$, 
We can re-write each $\Delta \dot{\bs{x}}_{i}$ using
the kinematics
% Given the end-effector velocity jump \eqref{eq:delta_ee_v}, 
\begin{equation}
\label{eq:fixed_coe}
\Delta \dot{\bs{x}}_{i} = J_{i} \Delta \dot{\bs{q}} \quad \text{for}~i = 1\ldots m+1,
\end{equation}
which simplifies \eqref{eq:impact_dynamics_general_middle} to: 
\begin{equation}
  \label{eq:impact_dynamics_general}
J\Delta \dot{\bs{q}}
=   
\Lambda^{-1} I. 
\end{equation}

Knowing the end-effector velocity jump $\Delta \dot{\bs{x}}_{m+1}$ from \eqref{eq:delta_ee_v}, 
we can predict  $\Delta \dot{\bs{q}}$, the impulse $I_{m+1}$ and the propagated impulses of the end-effectors with established contact $I_{i}$ for $i=1,\ldots,m$, using 
an auxiliary QP with the optimization variables $\bs{u} = [\Delta \dot{\bs{q}},  I_1, \ldots,  I_{m+1} ]^\top$:
\begin{equation}
\label{eq:impact_qp}
\begin{aligned}
\underset{\bs{u}} {\text{min}}\quad
&\frac{1}{2}\bs{u}^\top \bs{u}  \\
 \text{s.t.}\quad 
&\text{ Impulse propagation: \eqref{eq:impact_dynamics_general}}\\
&\text{ Initial condition:}~J_{m+1} \Delta \dot{\bs{q}} = \Delta \dot{\bs{x}}_{m+1}
\end{aligned}.
\end{equation}

% \begin{align*}
% \underset{\boldsymbol{u}} {\text{min}}\quad
% &\frac{1}{2}\boldsymbol{u}^\top \boldsymbol{u}  \\
%  \text{s.t.}\quad 
% &\text{ Impulse propagation:} \quad J\Delta \dot{\boldsymbol{q}}=\Lambda^{-1} I. \\
% &\text{ Initial condition:} \quad J_{m+1} \Delta \dot{\boldsymbol{q}} = \Delta \dot{\bs{x}}_{m+1}
% \end{align*}.

Since \eqref{eq:impact_qp} is an equality-constrained QP, its analytical solution is available.   Re-writing \eqref{eq:impact_qp} in the standard form:
\begin{equation}
\label{eq:least_norm_problem}
\begin{aligned}
\underset{\bs{u}}{\text{min}}\quad
&\frac{1}{2}\bs{u}^\top \bs{u}  \\
 \text{s.t.}\quad 
 &   [ J, ~-\Lambda^{-1}]\bs{u}  =   0,  \\
 & [ J_{m+1}, \quad 0~]\bs{u} = \Delta \dot{\bs{x}}_{m+1} 
\end{aligned}.
\end{equation}
The \emph{KKT system} associated with \eqref{eq:least_norm_problem} is: 
% \begin{equation*}
% \label{eq:kkt_system}
% \underbrace{\begin{bmatrix}
% I & A^\top\\
% A & 0
% \end{bmatrix}}_{K} \begin{bmatrix}
% \bs{u}\\
% \bs{\lambda}
% \end{bmatrix} = 
% \begin{bmatrix}
% 0\\
% \bs{b}
% \end{bmatrix},
% \end{equation*}
$$
\underbrace{\begin{bmatrix}
I & A^\top\\
A & 0
\end{bmatrix}}_{K} \begin{bmatrix}
\bs{u}\\
\bs{\lambda}
\end{bmatrix} = 
\begin{bmatrix}
0\\
\bs{b}
\end{bmatrix},
$$
where $\bs{\lambda}$ denotes the associated Lagrange multipliers,  $\bs{b} = [0,\ldots,\Delta \dot{\bs{x}}^\top_{m+1} ]^\top$ and $A = \begin{bmatrix}
J, &- \Lambda^{-1}\\
J_{m+1}, & 0
\end{bmatrix}. $
% \begin{equation}
% \label{eq:A_matrix}
% A = \begin{bmatrix}
% J, &- \Lambda^{-1}\\
% J_{m+1}, & 0
% \end{bmatrix}. 
% \end{equation}

As $\bs{b} \in \mathbb{R}^{3(m+1)}$ has all zeros except the last three elements, which is the predicted $\Delta \dot{\bs{x}}_{m+1}$ given by \eqref{eq:delta_ee_v}, we can predict the following for $i=1,\dots,m+1$ at time $t_{k+1}$:
\begin{equation}
\label{eq:optimal_solution}
\Delta \dot{\bs{q}}^* = K_{\Delta \dot{\bs{q}}}^{-1}\Delta \dot{\bs{x}}_{m+1}, I^{*}_i = K_i^{-1}\Delta \dot{\bs{x}}_{m+1},
\end{equation}
where $ K^{-1}_{\Delta \dot{\bs{q}}} \in \mathbb{R}^{n \times 3}$ and $ K^{-1}_{i} \in \mathbb{R}^{3 \times 3}$ are taken accordingly from the last three columns of the inverse $K^{-1}$.

The predictions defined in~\eqref{eq:optimal_solution} are  functions of $\ddot{\bs{q}}$ due to the predicted $\delta \dot{\bs{x}}_{m+1}$ \eqref{eq:delta_ee_v}. Thus we can use~\eqref{eq:optimal_solution} to formulate impact-aware constraints for a QP controller, e.g.~\eqref{eq:humanoids_qp}, to generate feasible motion in view of the hardware limits, existing unilateral contacts and ZMP conditions.

% \begin{remark}
% Due to the multiple contacts associated with \eqref{eq:impact_dynamics_general}, 
% we can find from the $k$th 
% row of  \eqref{eq:impact_dynamics_general}:
% $$
% \Delta \dot{\bs{x}}_k = \sum^{m+1}_{i=1}\Lambda^{-1}_{ki}I_i,
% $$ 
% which 
% indicates that the jump of one end-effector velocity $\Delta \dot{\bs{x}}_k$
% depends on all the other end-effectors with established contacts or under impacts. Thus
% the complementarity condition:
% \begin{equation*}
% % \label{eq:condition}
% \Delta \dot{\bs{x}}_i \geq 0, \quad I_i \geq 0, \quad \Delta \dot{\bs{x}}_i^{\top}I_i = 0.
% \end{equation*}
% does not hold for $\Delta \dot{\bs{x}}_i$ and $I_i$.
% \backfill
% \end{remark}
\begin{remark}
The least norm problem \eqref{eq:least_norm_problem} has a unique optimal solution $\bs{u}^{*} = K^{-1}\bs{b}$ as long as matrix 
$A$  has full row rank 
and $I$ is positive definite~\cite{boyd2004convex}. 
In view of  the components of matrix $A$, as long as the robot is not in a singular configuration, the conditions are fulfilled. 

If there is more than one impact, we can and add it to the auxiliary QP\eqref{eq:impact_qp} as an additional constraint $$J_{m+2} \delta \dot{\bs{q}} = \delta \dot{\bs{x}}_{m+2}.$$
\end{remark}

\subsection{Hardware limit constraints}
\label{sec:impact_hardware}
We formulate the constraints \eqref{eq:joint_velocity_impact_constraints} and \eqref{eq:joint_torque_impact_constraints} 
to prevent violating the hardware limits, i.e. the  
limited joint velocities
$[\underline{\dot{\bs{q}}}, \bar{\dot{\bs{q}}}]$ and the limited impulsive joint torques $[\underline{\bs{\tau}}, \bar{\bs{\tau}}]$. 
\subsubsection{Joint velocity limit}
As analyzed by \cite{wang2019rss}, we can restrict the post-impact joint velocity $\dot{\bs{q}}^{+} \in [\underline{\dot{\bs{q}}}, \bar{\dot{\bs{q}}}]$ by:
% We use the \emph{Euler forward} derivative approximation:
% \begin{equation*}
% \label{eq:euler_forward}
% \dot{\bs{q}}(t_k) = \frac{\bs{q}(t_{k+1}) - \bs{q}(t_{k}) }{\Delta t}
% \end{equation*}
% to incorporate the
% predicted $\Delta \dot{\bs{q}}^*(t_{k+1})$. 
% Using the definition of
% pre- and post-impact 
% velocity, we can expand the joint velocity limit constraint: 
% \begin{equation*}
% \resizebox{.99\hsize}{!}{$
% \begin{aligned}
%   -\ddot{\bs{q}}(t_k)    &\leq \frac{\dot{\bar{\bs{q}}} - \dot{\bs{q}}^{+}(t_{k+1})   }{\Delta t }  
% = \frac{\dot{\bar{\bs{q}}} - \dot{\bs{q}}(t_k) - \ddot{\bs{q}}(t_{k})\Delta t  - \Delta \dot{\bs{q}}(t_{k+1})   }{\Delta t }  \\
%    \ddot{\bs{q}}(t_k)    &\leq  - \frac{\underline{\dot{\bs{q}}} - \dot{\bs{q}}^{+}(t_{k+1})   }{\Delta t } 
%  = - \frac{\underline{\dot{\bs{q}}} - \dot{\bs{q}}(t_k)  - \ddot{\bs{q}}(t_{k})\Delta t - \Delta \dot{\bs{q}}(t_{k+1})}{\Delta t } 
% \end{aligned}
% $}
% \end{equation*}
% which can be simplified as: 
\begin{equation*}
\label{eq:delta_dq_ini}
\begin{aligned}
  \Delta \dot{\bs{q}}(t_{k+1})  &\leq  \dot{\bar{\bs{q}}} - \dot{\bs{q}}(t_k)   \\
- \Delta \dot{\bs{q}}(t_{k+1})  &\leq    - \left( \underline{\dot{\bs{q}}} - \dot{\bs{q}}(t_k)\right)
\end{aligned}.
\end{equation*}
We can reformulate the above to restrict $\ddot{\bs{q}}$: 
\begin{equation}
\label{eq:joint_velocity_impact_constraints}
\begin{aligned}
\mathcal{J}_{\Delta \dot{\bs{q}}}\ddot{\bs{q}} \Delta t &\leq \dot{\bar{\bs{q}}} - \dot{\bs{q}} -  \mathcal{J}_{\Delta \dot{\bs{q}}}\dot{\bs{q}}   \\
- \mathcal{J}_{\Delta \dot{\bs{q}}}\ddot{\bs{q}} \Delta t &\leq - \left( \dot{\underline{\bs{q}}} - \dot{\bs{q}} -  \mathcal{J}_{\Delta \dot{\bs{q}}}\dot{\bs{q}} \right)
\end{aligned},
\end{equation}
where $\mathcal{J}_{\Delta \dot{\bs{q}}}$ is defined in view of
the predicted
$\Delta \dot{\bs{x}}_{m+1}$ \eqref{eq:delta_ee_v} and  $\Delta \dot{\bs{q}}$
\eqref{eq:optimal_solution}: 
\begin{equation*}
\Delta \dot{\bs{q}}(t_{k+1}) = \underbrace{
  K_{\Delta \dot{\bs{q}}}^{-1}P_{\Delta}J_{m+1}
}_
{\mathcal{J}_{\Delta \dot{\bs{q}} } }
\left(\dot{\bs{q}}(t_{k}) + \ddot{\bs{q}}(t_{k})\Delta t \right).
\end{equation*}
\subsubsection{Impulsive joint torque}
Following previous examples, e.g. \cite{konno2011impact},
we  define the impulsive end-effector forces:
$$
\bar{\bs{f}}_i = \frac{I_i}{\delta t}\quad \text{for}~i = 1, \ldots m + 1.
$$ 
We can  predict the whole-body impulsive joint torque:
\begin{equation*}
\label{eq:impulsive_tau}
\Delta \bs{\tau} = \sum^{m+1}_{i=1} \bs{\tau}_i  = \sum^{m+1}_{i=1} J_i^{\top}\bar{\bs{f}}_i 
= \frac{1}{\delta t }(\sum^{m+1}_{i=1} J_i^{\top} K_i^{-1}) \Delta \dot{\bs{x}}_{m+1},
\end{equation*}
and restrict it by: 
% In case of the impulsive joint torque constraint $\Delta \underline{\bs{\tau}} \leq \Delta \bs{\tau} \leq \Delta \bar{\bs{\tau}}$, we can directly reformulate it as: 

\begin{equation}
\label{eq:joint_torque_impact_constraints}
\begin{aligned}
 \frac{\Delta t}{\delta t} \mathcal{J}_{\Delta \bs{\tau}}\ddot{\bs{q}} 
&\leq 
\Delta \bar{\bs{\tau}} - \frac{1}{\delta t} \mathcal{J}_{\Delta \bs{\tau}} \dot{\bs{q}}\\
 - \frac{\Delta t}{\delta t} \mathcal{J}_{\Delta \bs{\tau}}\ddot{\bs{q}} 
&\leq 
- ( \Delta \underline{\bs{\tau}} - \frac{1}{\delta t} \mathcal{J}_{\Delta \bs{\tau}} \dot{\bs{q}})
\end{aligned},
\end{equation}
where $\mathcal{J}_{\Delta \bs{\tau}}$ is defined using
the predicted 
$\Delta \dot{\bs{x}}_{m+1}$  and  $\Delta \bs{\tau}$:
$$
\Delta \bs{\tau}(t_{k+1}) = 
\frac{1}{\delta t}\underbrace{ 
(\sum^{m+1}_{i=1} J_i^{\top} K_i^{-1})
P_{\Delta}J_{m+1}
}_
{\mathcal{J}_{\Delta \bs{\tau}} }
\left(\dot{\bs{q}}(t_{k}) + \ddot{\bs{q}}(t_{k})\Delta t \right).
$$

\subsection{Holding Established Contacts}
\label{sec:impact_contact_maintenance}
We propose the constraint \eqref{eq:post_impact_geometric_constraints} and \eqref{eq:post_impact_slippage_constraints} to sustain an established contact by  restricting  the center of pressure and fulfilling the friction cone. 

\subsubsection{Center of pressure constraint}
\label{sec:impact_contact_maintenance_geometric}
Due to the propagated impulse, the constraint \eqref{eq:CoP_constriants} becomes
% We  relax 
% the equality constraint \eqref{eq:post_impact_geometry_dynamics} to the following inequality:
\begin{equation*}
A_{c}( \bar{\bs{F}} + \bs{F}) \leq 0
 \Rightarrow A_{c}\bar{\bs{F}} \leq - A_{c}\bs{F},
\end{equation*}
where $\bs{F}$ denotes the measured wrench of an established contact and $\bar{\bs{F}} = [\bs{0}^{\top}, \bar{\bs{f}}^{\top}]$. Let $A_{c2}$ as the  columns of $A_c$ corresponding to force, we have: 
$$
A_{c2}
\bar{\bs{f}}
 \leq - A_{c}\bs{F}.
$$
We can re-write the above to restrict $\ddot{\bs{q}}$:
\begin{equation}
\label{eq:post_impact_geometric_constraints}
A_{c2}\mathcal{J}_{f}\ddot{\bs{q}} \frac{\Delta t}{\delta t}
 \leq 
- A_{c} \bs{F} - A_{c2}\mathcal{J}_{f}\dot{\bs{q}}\frac{1}{\delta t},
\end{equation}
where we defined the Jacobian $\mathcal{J}_{f}$ in view of the impulsive force and the predicted impulse $I^*$ \eqref{eq:optimal_solution}:
$$
\bar{\bs{f}}  = \frac{I^*}{\delta t}  =  \frac{1}{\delta t}
\underbrace{
K_i^{-1}
P_{\Delta}J_{m+1}
}_
{\mathcal{J}_{f} }
\left(\dot{\bs{q}}(t_{k}) + \ddot{\bs{q}}(t_{k})\Delta t \right).
$$

\subsubsection{Fulfilling friction cone}
We prevent slippage by limiting the predicted contact force within the friction cone: 
\begin{equation*}
N_{\vec{\bs{n}}}(\bs{f} + \bar{\bs{f}}) \leq \mu P_{\vec{\bs{n}}}(\bs{f} + \bar{\bs{f}}). 
\end{equation*}
If we define $P_{\mu} = I - \vec{\bs{n}}\vec{\bs{n}}^\top - \mu\vec{\bs{n}}\vec{\bs{n}}^\top$, we can re-write the constraint as $P_{\mu}\bar{\bs{f}}
 \leq 
- P_{\mu}\bs{f}
$ 
or equivalently:
\begin{equation}
\label{eq:post_impact_slippage_constraints}
P_{\mu}\mathcal{J}_{f}\ddot{\bs{q}} \frac{\Delta t}{\delta t}
 \leq 
- P_{\mu} (\bs{f} + \mathcal{J}_{f}\dot{\bs{q}}\frac{1}{\delta t}).
\end{equation}

\subsection{Bounded ZMP}
\label{sec:impact_ZMP}
Given the impulsive forces of all
the end-effectors either with an established contact or undergoing an impact, 
we can predict the impact-induced jump of the ZMP to  
fulfill
the ZMP constraint \eqref{eq:ZMP_constraints}:
$$
A_{\text{Z}}(
\sum^{m+1}_{i =1}\bs{F} +  \sum^{m+1}_{i =1} \bar{\bs{F}}) \leq 0 \Rightarrow A_{\text{Z}} \sum^{m+1}_{i =1} A_{\text{i}}\bar{\bs{f}}_i
\leq  - \sum^{m+1}_{i =1} A_{\text{Z}}
\bs{F}_i,
$$
where $A_{\text{i}}$ denotes the transformation matrix that calculates the equivalent wrench in the
inertial frame due to the impulsive force $\bar{\bs{f}}_i$,
% denotes the first three columns of  
% the adjoint map $Ad^\top_{g_{i\text{0}}} \in \mathbb{R}^{6 \times 6}$ between the $i$-th contact frame and the inertial frame, 
the wrench $\bs{F}$ denotes the sum of the external wrenches in the inertial frame. Thus we
can restrict the robot joint accelerations $\ddot{\bs{q}}$ with the following inequality: 

\begin{equation}
\label{eq:post_impact_zmp_constraints}
A_{\text{Z}}  \sum^{m+1}_{i =1}(A_{\text{i}}\mathcal{J}_{f_i})\ddot{\bs{q}} \frac{\Delta t}{\delta t}
\leq  - A_{\text{Z}}(\sum^{m+1}_{i =1} \bs{F}_i + \frac{1}{\delta t} \sum^{m+1}_{i =1} (A_{\text{i}}\mathcal{J}_{f_i}) \dot{\bs{q}}).
\end{equation}

\subsection{Impact-robust QP controller synthesis}
\label{sec:impact_qp_controller}
Should there exists an incoming impact at one end-effector, we need to solve the modified QP: 
\begin{equation}
\label{eq:robust_qp_controller}
\begin{aligned}
  \min_{
\bs{x}: (\ddot{\bs{q}}, \bs{f}_{\lambda})} \quad & \sum_{i \in \mathcal{I}_{o}}  w_i \|\bs{e}_i (\bs{x}) \|^2   
  \\
  \mbox{s.t.}\quad & \text{Common usual constraints},\\
  \quad&\text{Hardware constraints:}  \eqref{eq:joint_velocity_impact_constraints} \eqref{eq:joint_torque_impact_constraints} , \\
   \quad& \text{Holding contact constraints:} \eqref{eq:post_impact_geometric_constraints}, \eqref{eq:post_impact_slippage_constraints} , \\
   \quad& \text{ZMP constraint:} \eqref{eq:post_impact_zmp_constraints}. 
\end{aligned}
\end{equation}

Compared to the usual QP controller  \eqref{eq:humanoids_qp},  
a humanoid robot controlled by \eqref{eq:robust_qp_controller} is always
able to guarantee the hardware limits, maintain the established contacts and the ZMP condition while fulfilling the task objectives included in $\mathcal{I}_{o}$. Thus we do not need to make any assumption on the impact timing or manually choose a safe yet near-zero contact velocity, 
rather, \eqref{eq:robust_qp_controller} would generate the maximal contact velocities with respect to the feasibility of the constraints (\ref{eq:joint_velocity_impact_constraints}-\ref{eq:post_impact_zmp_constraints}).

\section{Experiment}
\label{sec:experiments}

We use a full-size humanoid robot HRP-4 to validate that the proposed QP controller is able to generate feasible contact velocities with respect to the constraints (\ref{eq:joint_velocity_impact_constraints}-\ref{eq:post_impact_zmp_constraints}). Using the experimental parameters  summarized in Sec.~\ref{sec:parameters}, 
we present the experimental results 
in Sec.~\ref{sec:push-wall}
where the
robot generated the maximal contact velocity along the direction of interest  
rather than planning the contact at a specific location, i.e. being aware of the contact surface location. In the snapshots shown in Fig~\ref{fig:experiment-snapshots}, the robot hit the wall with a contact velocity of $0.35~m/s$ and then regulate the contact force to $15~N$.
% Then in Sec.~\ref{sec:grab-box}, the same set of constraints is used  in a box-grabbing experiment. Compared to conventional approaches where the robot needs to established the contact and then entered the following steps, the robot grabbed a box with swift motion.
We encourage interested readers to 
check the \href{https://youtu.be/FovQG6U448Q}{experiment videos}.
% and the implementation details\footnote{\url{https://github.com/wyqsnddd/mc_impact_pusher}}.

% Despite the robot has $40$~DoFs, 
% we generated  whole-body motion that leads to a
% contact velocity at $0.23~m/s$,
% see the snapshots in Fig~\ref{fig:experiment-snapshots}.

\subsection{Parameters}
\label{sec:parameters}
% Stiff motion is needed to exert
In order to exert the impulse,  
We mounted a 3D printed plastic palm of $3~$cm thickness. 
The 
robot keeps the maximal available 
contact velocity until 
the force sensor mounted on the wrist reached the impact detection threshold, i.e. $20$~N.
In order to correctly observe the post-impact state jumps, we choose not to use a stabilizer. 
For the established contacts, we choose the friction coefficient as $0.7$.

% to avoid propagating the impulse to the rigid yet fragile components, e.g. harmonic drives. 

% We do not need to make assumption of the exact location of the impact, i.e. we set up a whiteboard somewhere in front of the robot

% In order to avoid hardware damage, 
% we simulate the HRP-4 robot to punch a fixed wall at $0.6~m/s$ to 

% validate the constraints (\ref{eq:joint_velocity_impact_constraints}-\ref{eq:post_impact_zmp_constraints}) in Sec.~\ref{sec:constraints}. Then we 
% present the hardware experiments in Sec.~\ref{sec:experiment} where 
% the robot is able to lift up boxes at different weights using swift motions without stopping like a human.       

% The contact maintenance and balance constraints  (\ref{eq:post_impact_geometric_constraints}-\ref{eq:post_impact_zmp_constraints}) are more sensible to impacts, i.e. they are violated prior to the constraints (\ref{eq:joint_velocity_impact_constraints}-\ref{eq:joint_torque_impact_constraints}).

% Within the feasible search space provided by the 
% constraints  (\ref{eq:joint_velocity_impact_constraints}-\ref{eq:post_impact_zmp_constraints}), 

Based on several trial runs, we choose the coefficient of restitution $c_r =0.02$,  which indicates trivial rebounce. It leads to a reasonable prediction of the impulsive force, see  Fig.~\ref{fig:impact-force}.

The QP controller runs at $200~Hz$, which gives the sampling period $\Delta t = 5~ms$. The impact duration $\delta t$ appears in the constraints  (\ref{eq:joint_torque_impact_constraints}-\ref{eq:post_impact_zmp_constraints}),  where the predicted impulsive forces are used. As the ATI-45 force-torque sensors are read at $200~Hz$, we choose the same period for the impact duration: $\delta t = 5~ms$.

% \subsection{Coefficient of restitution estimation}

% Based on the joint space impact dynamics model and  
% the postive definiteness of the mass matrix $M(\bs{q})$, we can calculate the impact-induced joint velocity jump
% $$
% \delta \dot{\bs{q}} = M^{-1}J^{\top} \bar{\bs{f}}\delta t,
% $$ 
% which leads to the the end-effector velocity jump $\delta \dot{\bs{x}} = J \delta \dot{\bs{q}}$, see \cite{wang2019rss}. Given the average impulsive force $\bar{\bs{f}}$ measured by the ATI-mini45 force torque sensor, we can estimate the coefficent of restitution as: 
% \begin{equation*}
% \hat{c}_{\text{r}} = - \frac{\dot{\bs{x}}^{+}}{P_{\vec{\bs{n}}} \dot{\bs{x}}^{-}} = 
% - \frac{  \delta \dot{\bs{x}} +  P_{\vec{\bs{n}}} \dot{\bs{x}}^{-} }{P_{\vec{\bs{n}}} \dot{\bs{x}}^{-}} = 
% - \frac{ J M^{-1}J^{\top} \bar{\bs{f}}\delta t  +  P_{\vec{\bs{n}}} J \dot{\bs{q}} }{P_{\vec{\bs{n}}} J \dot{\bs{q}}}.
% \end{equation*}

\subsection{Constraints validation}
\label{sec:push-wall}
The proposed QP controller \eqref{eq:robust_qp_controller} autonomously determines the \emph{feasible} contact velocity. Thus 
we assign an exceptionally high contact velocity, i.e. $0.8~m/s$, to check if the hardware limits and the standing stability are satisfied. 
The ZMP profiles generated with different constraints settings reveal that the support polygon $\mathcal{S}$, which is the enclosing convex polygon of the feet contact areas, is  too conservative. More applicable stability measures or an extended support polygon are needed to exploit the maximal contact velocity.

% constraints (\ref{eq:joint_velocity_impact_constraints}-\ref{eq:post_impact_zmp_constraints}) are satisfied respectively. 

\begin{figure}[!htp]
 \centering
 \includegraphics[width=0.8\columnwidth, height=3.5cm]{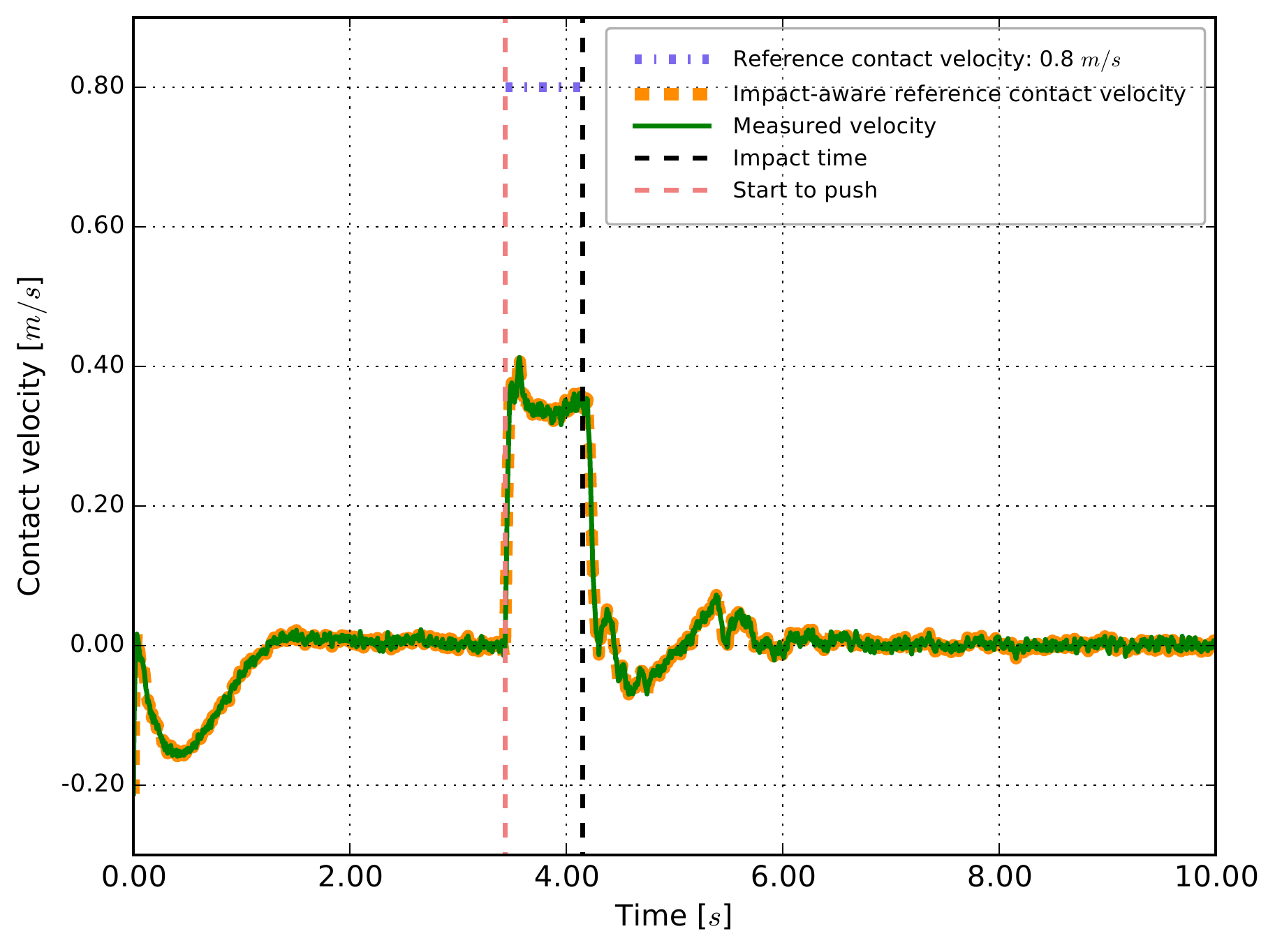}
 \caption{
Given the reference $0.8~m/s$, the QP controller \eqref{eq:robust_qp_controller} autonomously determined
the safe contact velocity. 
}
\label{fig:contact-vel}
\end{figure}

\begin{figure}[!htp]
 \centering
 \includegraphics[width=1.0\columnwidth, height=5.5cm]{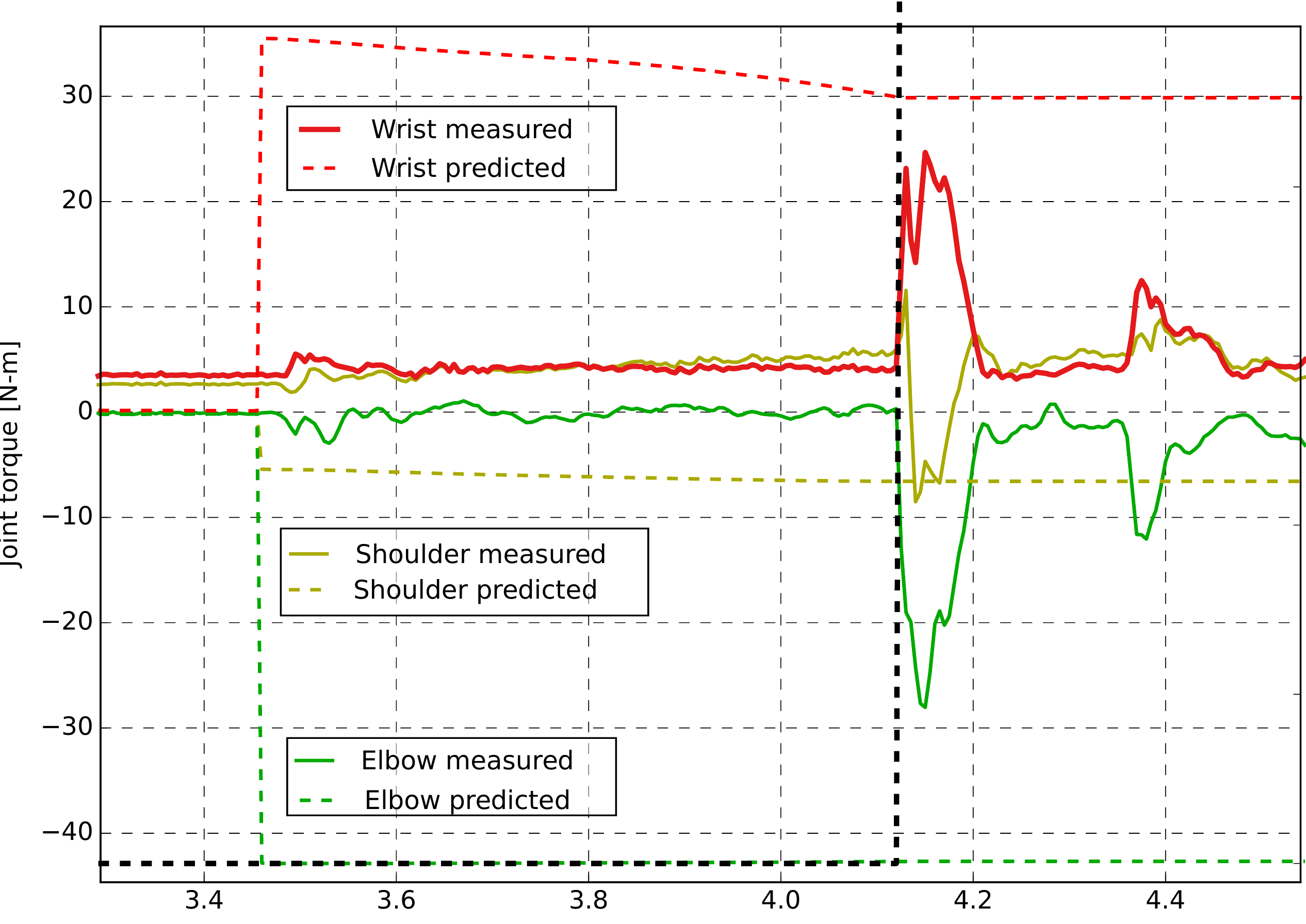}
 \caption{
   Considering  three joints taken from the shoulder, elbow, and wrist, 
   at the impact time the joint torque 
calculated by $J^{\top}\bs{f}$, where $\bs{f}$ is read from the sensor, is close to the prediction $\Delta \bs{\tau}$ and smaller than the corresponding bounds of $\pm46~N\cdot m$, $\pm42.85~N\cdot m$ and $\pm85.65~N\cdot m$.
}
\label{fig:impulsive-torque}
\vspace{-4mm}
\end{figure}

In order to fulfill the impulsive joint torque bounds, see Fig.~\ref{fig:impulsive-torque},
% According to the predicted impulsive force shown in Fig.~\ref{fig:impact-force}, 
the QP controller \eqref{eq:robust_qp_controller} updates the  \emph{feasible} contact velocity set-point in real-time as shown  in Fig.~\ref{fig:contact-vel}.
% which leads to the bounded impulsive joint torque, see Fig.~\ref{fig:impulsive-torque}.
In all the plots, we use a dashed black line to indicate the impact time.


\begin{figure}[!htp]
 \centering
\includegraphics[width=1.0\columnwidth]{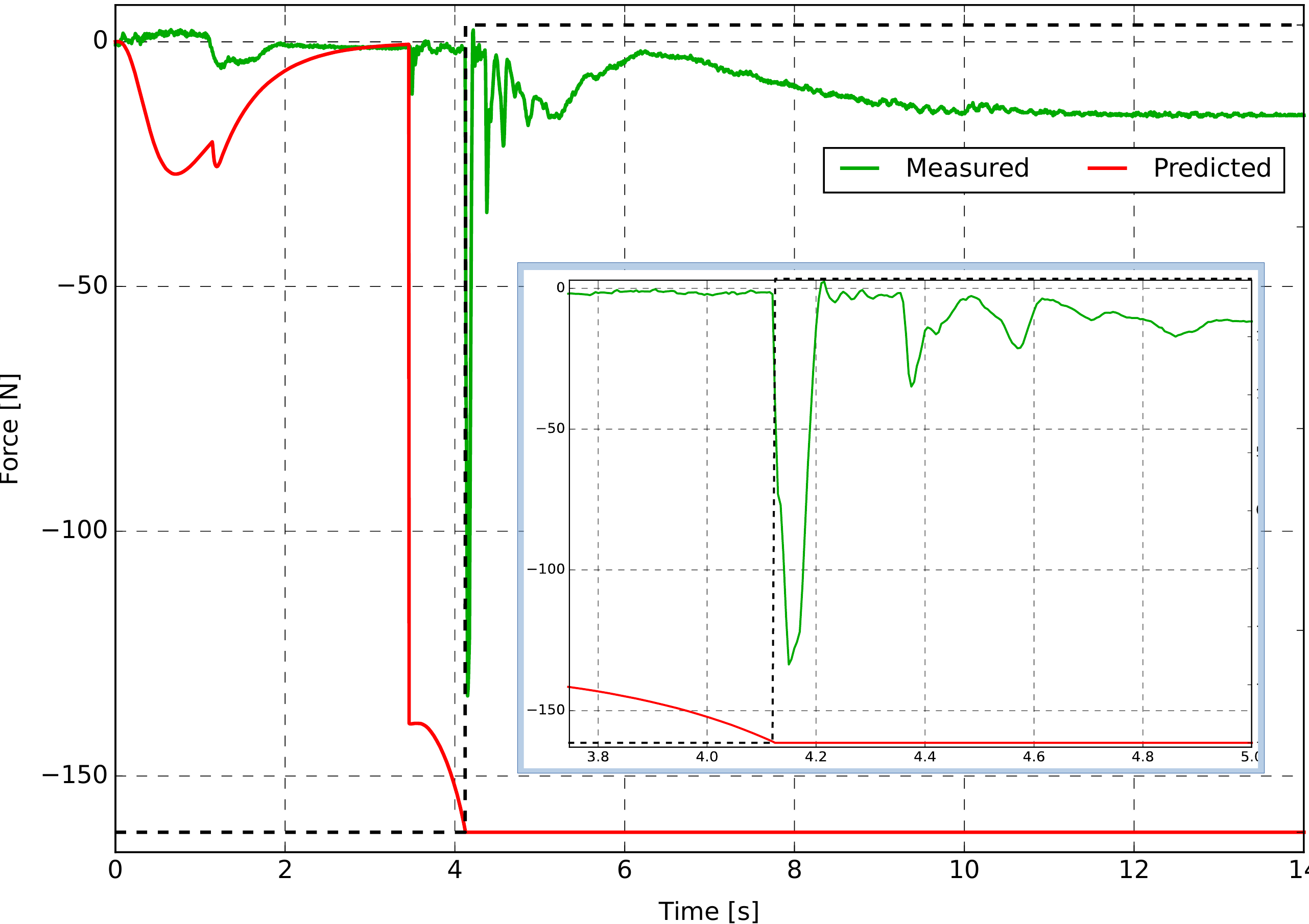}
 \caption{
%    The predicted impulsive force 
% $\bar{f} = \frac{I}{\delta t}$, where the impulse $I$ is calculated by solving
% the auxiliary QP \eqref{eq:least_norm_problem} and the impact duration is chosen to be $0.005~s$. 
The impact detected at $4.18~s$ generated 
impulsive force $133~N$ which is smaller the predicted impulsive force $161~N$. The conservative prediction  leads to safe motion generation in view of the worst-case impact. 
}
\label{fig:impact-force}
\vspace{-5mm}
\end{figure}

The predicted impulsive force is shown in Fig.~\ref{fig:impact-force}. After the impact is detected, 
the contact velocity did not reduce to zero
as 
the robot started an admittance controller to regulate the \emph{post-impact} contact force to $15~N$, which is shown between $10$s and $15$s of Fig.~\ref{fig:impact-force}.
% \begin{figure}[!htp]
%  \centering
% \includegraphics[width=1.0\columnwidth, height=5cm]{figure/experiment-aug-29/no-all-force/post-impact-admittance-crop-new}
%  \caption{
% The post-impact contact force is regulated to $15~N$.
% }
% \label{fig:admittance-task}
% \vspace{-6mm}
% \end{figure}

\begin{figure}[!htp]
 \centering
 \includegraphics[width=0.9\columnwidth]{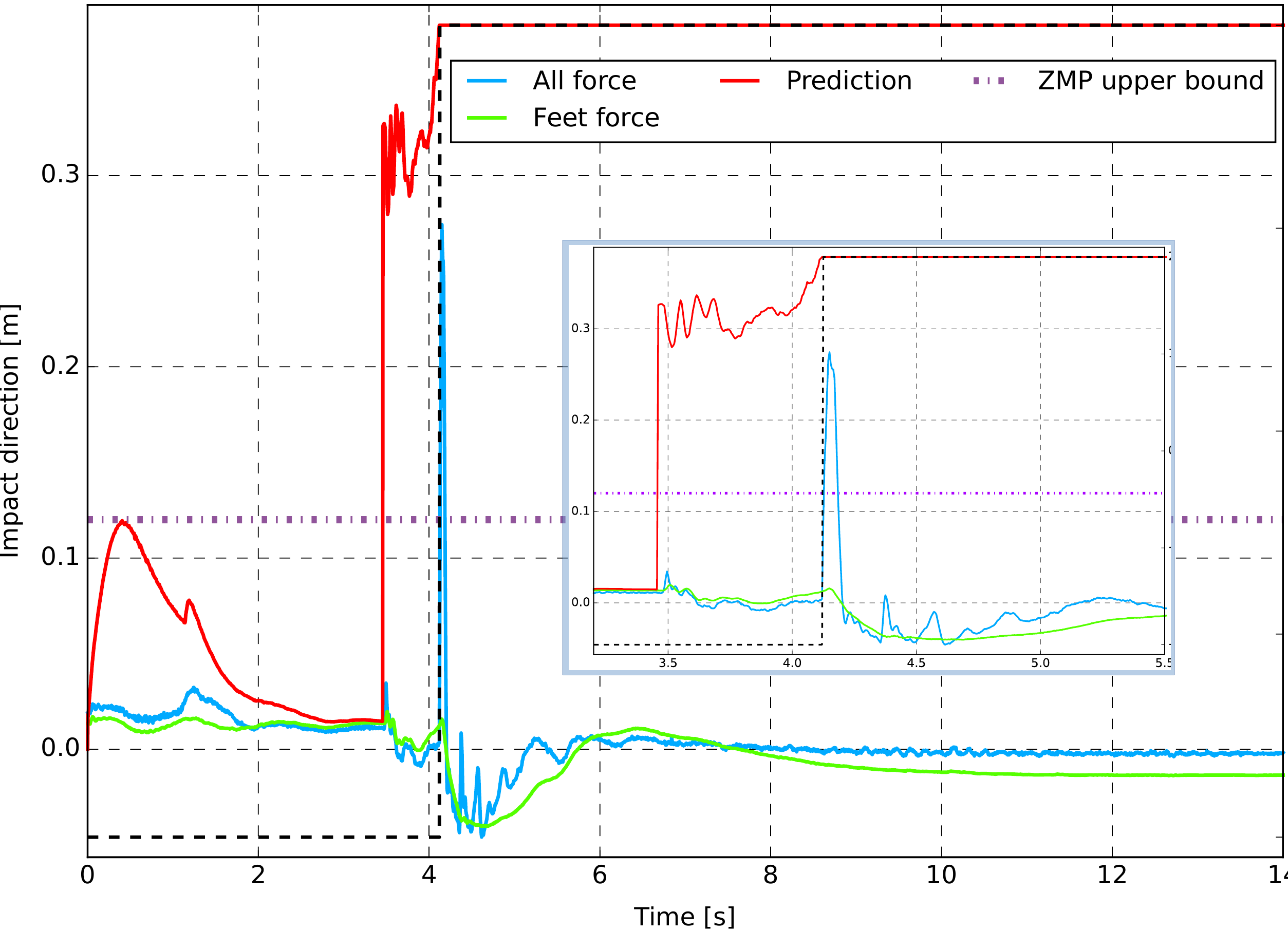}
 \caption{
Comparison of ZMP (along the impact direction) computed with feet force (light green), both feet and measured impulsive force at the wrist (light blue), and both feet and predicted impulsive force at the wrist (red).
}
\label{fig:zmp-comparison-x}
\vspace{-3mm}
\end{figure}

\begin{figure}[t!]
    \centering
    \begin{subfigure}[t]{0.25\textwidth}
    % \begin{subfigure}
      \centering
      \includegraphics[height=4.5cm]{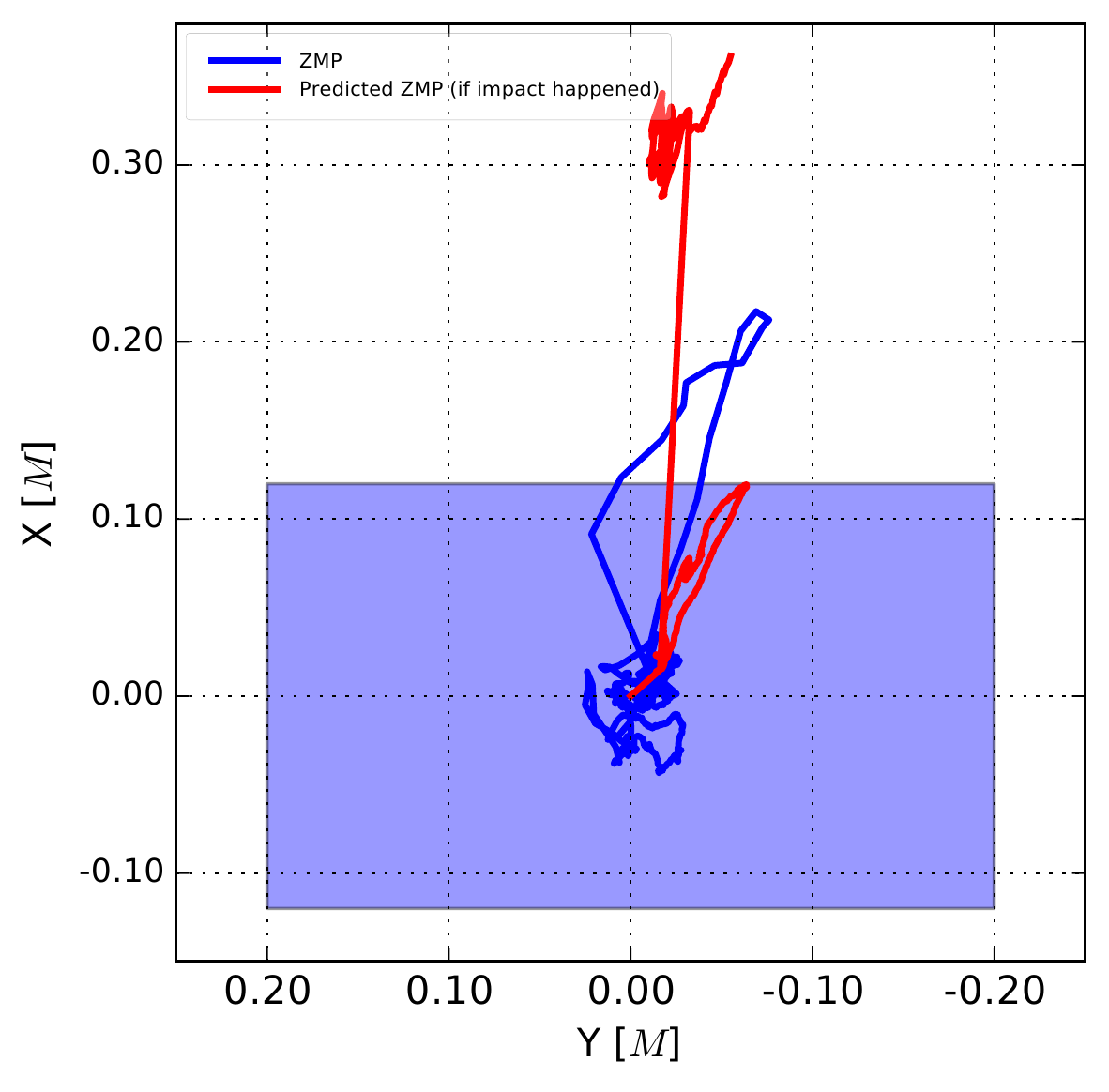}
      \caption{Contact velocity $0.35m/s$.}
      \label{fig:zmp-lost}
    \end{subfigure}%
    ~ 
    \begin{subfigure}[t]{0.25\textwidth}
    % \begin{subfigure}
      \centering
       \includegraphics[height=2.8cm]{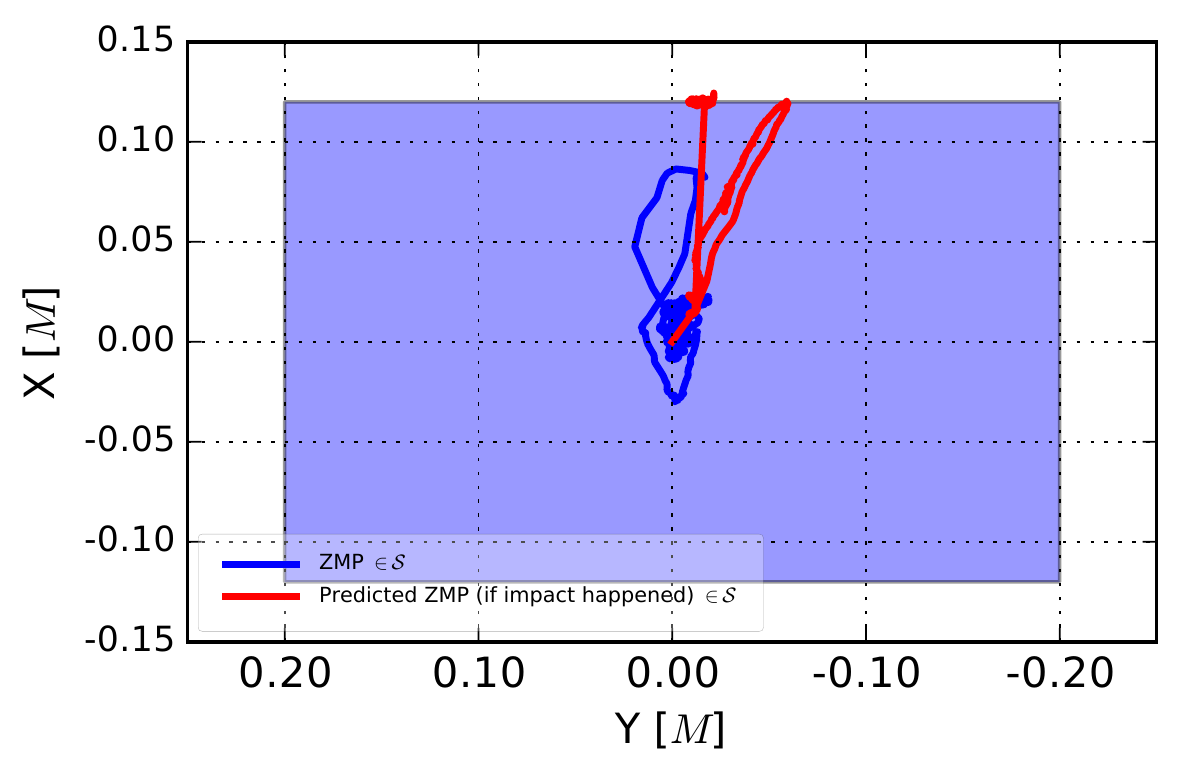}
       \caption{Contact velocity $0.11m/s$.}
       \label{fig:ZMP-strict}
    \end{subfigure}
    \caption{
In fig.~\ref{fig:zmp-lost} we can predict the ZMP would jump outside the support polygon $\mathcal{S}$ due to
the contact velocity $0.35~m/s$. In fig.~\ref{fig:ZMP-strict}  the contact velocity is reduced to $0.11~m/s$,  the ZMP is strictly bounded within the support polygon $\mathcal{S}$. 
    }
\end{figure}

From the snapshots in Fig.~\ref{fig:experiment-snapshots}, the robot did not fall. However if we check the ZMP (along the normal direction of the wall surface) plotted in Fig.~\ref{fig:zmp-comparison-x}, we can find that the ZMP temporarily jumped outside the bound (see the 2D view in Fig.~\ref{fig:zmp-lost}). 
% calculated by including the measured impact force of the hand (light blue curve)
% despite that the ZMP calculated  merely using the feet force measurements (light green curve) is within the bound all the time. 

% \begin{figure}[!htp]
%  \centering
%  \includegraphics[width=0.6\columnwidth, height=4.5cm]{figure/ZMP-lost-new.pdf}
%  \caption{
% We can predict the ZMP would jump outside the support polygon $\mathcal{S}$ due to
% the contact velocity $0.35~m/s$.
% }
% % \label{fig:zmp-lost}
% \vspace{-5mm}
% \end{figure}

\begin{figure}[!htp]
 \centering
 \includegraphics[width=0.9\columnwidth]{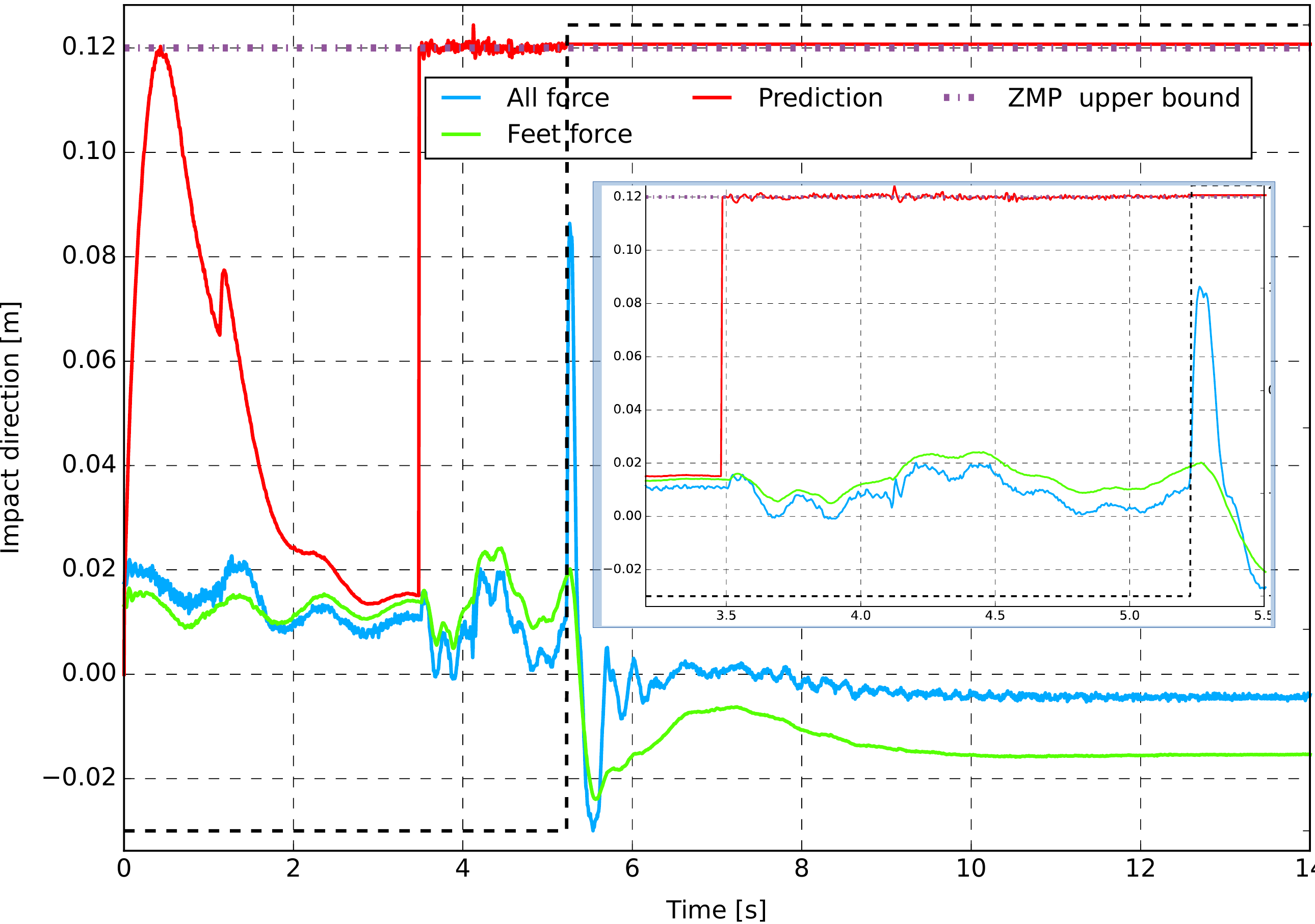}
 \caption{
Using predicted impulsive force at the wrist in constraint
\eqref{eq:post_impact_zmp_constraints}, we can strictly bound the ZMP $\bs{z} \in \mathcal{S}$.
}
\label{fig:zmp-comparison-x-allforce}
\vspace{-5mm}
\end{figure}

Using the predicted impulsive force shown in Fig.~\ref{fig:impact-force}, we can actually predict the
ZMP jump.
In Fig.~\ref{fig:zmp-comparison-x}, we can see that the predicted ZMP in case of impact (red curve) is well above the actual jump. Thus using this information we can keep the ZMP strictly bounded.
In Fig.~\ref{fig:zmp-comparison-x-allforce} and the 2D view Fig.~\ref{fig:ZMP-strict}, we plot the ZMP of another experiment where the only difference is that the wrench generated by the predicted impulsive force of the hand is included in the constraint \eqref{eq:post_impact_zmp_constraints}. Compared to Fig.~\ref{fig:zmp-comparison-x} and Fig.~\ref{fig:zmp-lost}, both
the predicted ZMP under impact and the actual ZMP  are well bounded by the support polygon. Not surprisingly, the price we paid for being more stable is  slowing down the contact velocity to $0.11~m/s$.
% , see Fig.~\ref{fig:contact-vel-allforce}.
\section{Conclusion}

Impact-induced state jumps, i.e. joint velocity and impulsive forces, challenge the hardware limits of a robot. In the case of a humanoid, the problem gets even more complicated due to its
complicated kinematic structure and additional requirements for 
contact  and balance maintenance.
Through analysis of the impact-induced state jumps propagation between different kinematic branches, we propose a set of modified constraints to guarantee the feasibility of the robot configuration such that  
the QP controller can exploit the maximal contact velocity of a humanoid robot. 
Through experiments performed by an HRP-4 robot, we achieved contact velocity at $0.35~m/s$ and maximal impulsive force $133~N$, which are significant compared to the motion generated by the conventional impedance control law.
% First 
To the best of our knowledge, we are the first to propose an impact-aware humanoid robot motion generation controller based on quadratic optimization. 

In the future, we need a less conservative stability condition rather than restricting ZMP strictly inside the support polygon: $\bs{z} \in \mathcal{S}$, whose conservativeness has been already revealed from the experiments. 

\bibliography{ref}
\bibliographystyle{IEEEtran}

\end{document}